\newif\ifsupp  
\supptrue

\documentclass{article}

\usepackage{fullpage}

\usepackage{amsthm,amsmath}
\usepackage{bbm}
\usepackage{times}
\usepackage{array,float}
\usepackage{url}
\usepackage{amstext,amssymb}
\usepackage{hyphenat}
\usepackage{verbatim}
\usepackage{dsfont}
\usepackage{bm}
\usepackage{wrapfig}
\usepackage[noend]{algorithmic}
\usepackage{algorithm}
\usepackage{graphicx,color}
\usepackage{xparse,etoolbox}
\usepackage{threeparttable}
\usepackage{dsfont}
\usepackage{xspace}
\usepackage{subcaption}
\usepackage[utf8]{inputenc}
\usepackage{outlines}
\usepackage{comment}
\usepackage{footnote}
\usepackage{multirow}
\usepackage{ulem}
\usepackage{hyperref}
\usepackage{cleveref}
\usepackage{authblk}

\newtheorem{lemma}{Lemma}[section]

\newtheorem{defn}[lemma]{Definition}

\newcommand{\SUB}[1]{\hspace{-0.15in} \textbf{#1}}

\newcommand{\moments}{\mathcal{M}}
\newcommand{\sample}{\mathcal{C}}
\newcommand{\update}{\Delta}
\newcommand{\Normal}{\mathcal{N}}
\newcommand{\lepochs}{\ensuremath{E}}

\newcommand{\lbs}{\ensuremath{B}}
\newcommand{\loss}{\ell}
\newcommand{\mycaptionof}[2]{\captionof{#1}{#2}}

\renewcommand{\paragraph}[1]{\vspace{3pt}\noindent\textbf{#1}}

\newcommand{\norm}[1]{\|#1\|}

\newcommand{\eps}{\epsilon}

\newcommand{\cA}{\mathcal{A}}

\newcommand{\cS}{\mathcal{S}}

\newcommand{\R}{\mathcal{R}}

\newcommand{\grad}{\bigtriangledown}
\newcommand{\mypar}[1]{\bigskip
	\noindent{\textbf{\em {#1}:}}}

\newcommand{\ignore}[1]{}

\makeatletter
\def\blfootnote{\gdef\@thefnmark{}\@footnotetext}
\makeatother

\makeatletter
\newcommand{\skipitems}[1]{\addtocounter{\@enumctr}{#1}}
\makeatother

\renewcommand{\paragraph}[1]{\medskip
\noindent{\bf{#1}}}

\title{Understanding Unintended Memorization \\ in Federated Learning}
\author{Om Thakkar}
\author{Swaroop Ramaswamy}
\author{Rajiv Mathews}
\author{Françoise Beaufays}
\affil{Google LLC,\\
Mountain View, CA, U.S.A. \\
\texttt{\{omthkkr, swaroopram, mathews, fsb\}   @google.com}}

\begin{document}

\maketitle

\begin{abstract}
Recent works have shown that generative sequence models (e.g., language models) have a tendency to memorize rare or unique sequences in the training data. Since useful models are often trained on sensitive data, to ensure the privacy of the training data it is critical to identify and mitigate such \emph{unintended} memorization. Federated Learning (FL) has emerged as a novel framework for large-scale distributed learning tasks. However, it differs in many aspects from the well-studied \emph{central learning} setting where all the data is stored at the central server. In this paper, we initiate a formal study to understand the effect of different components of canonical FL on unintended memorization in trained models, comparing with the central learning setting. Our results show that several differing components of FL play an important role in reducing unintended memorization. Specifically, we observe that the clustering of data according to users---which happens by design in FL---has a significant effect in reducing such memorization, and using the method of Federated Averaging for training causes a further reduction. We also show that training with a strong user-level differential privacy guarantee results in models that exhibit the least amount of unintended memorization.
\end{abstract}

\section{Introduction}
\label{sec:intro}

There is a growing line of work \cite{mod-inv1, mod-inv2,mem-inf1,CLKES18,SS19} demonstrating that neural networks can leak information about the underlying training data in unexpected ways. In particular, many of these works show that generative sequence models, which include commonly-used language models, are prone to \emph{unintentionally memorize} rarely-occurring phrases in the data. Large-scale learning often involves training over sensitive data, and such memorization can result in blatant leaks of privacy (e.g., \cite{xkcd}). Thus, for any novel learning framework of interest, it is crucial to test the resilience of models trained in the framework against such memorization. Techniques to mitigate memorization must be identified to ensure the privacy of training data.

The framework of Federated Learning (FL)~\cite{FL1, FL4} has emerged as a popular approach for training neural networks on a large corpus of decentralized on-device data (e.g., \cite{FL2, FL3, FL5, NWP18, FL19}).
FL operates in an iterative fashion: in each round, sampled client devices receive the current global model from a central server to compute an update on their locally-stored data, and the server aggregates these updates using the Federated Averaging algorithm~\cite{FL1} to build a new global model.
A hallmark of FL is that each participating device only sends model weights to the central server; raw data never leaves the device, remaining locally-cached. Although this, by itself, is not sufficient to provide formal privacy guarantees  for the training data, it is important to note that the canonical setting of FL~\cite{FL1} does differ in many aspects from the well-studied \emph{central learning} setting where all the data is stored at a central server. In this work, we initiate a formal study to understand the effect of the different components of FL, compared to the central learning setting, on unintended memorization in trained models.

We also investigate the effect of using a training procedure, with a formalized privacy guarantee, on such memorization. To this end, we use Differential Privacy (DP) ~\cite{DMNS06, DKMMN06}, which has become the standard for performing learning tasks over sensitive data. DP has been adopted by companies like Google~\cite{rappor, esa, esa2}, Apple~\cite{apple_report}, Microsoft~\cite{mic}, and LinkedIn~\cite{link}, as well as the US Census Bureau~\cite{census2}. Intuitively, DP prevents an adversary from confidently making any conclusions about whether any particular user's data was to train a model, even while having access to the model and arbitrary external side information.

 \begin{figure}[ht]
	\centering
	\begin{subfigure}[b]{0.49 \textwidth}
	    \centering
		\includegraphics[width=0.7\linewidth]{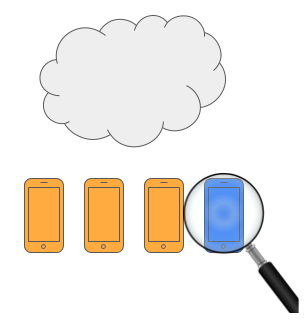}
		\caption{A user being selected as a secret sharer for a canary.}
		\label{fig:p_u}
	\end{subfigure}
	\qquad
	\begin{subfigure}[b]{0.4 \textwidth}
	    \centering
		\includegraphics[width=0.82 \linewidth]{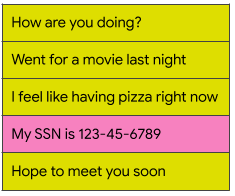}
		\caption{An example in a secret sharer's dataset being replaced by the canary.}
		\label{fig:p_e}
	\end{subfigure}
    \caption{An illustration of our federated secret-sharer framework, using the canary phrase ``My SSN is 123-45-6789" with a user-selection probability $p_u=0.25$, and an example-replacement probability $p_e=0.2$. Thus, in expectation, 1 out of every 4 users will be a secret sharer for this canary, and 1 out of every 5 examples in each such secret sharer's data will be replaced by the canary.}
	\label{fig:fss}
\end{figure}

 We build on the ``secret sharer" framework from \cite{CLKES18} that was designed to measure the unintended memorization in generative models.
 At a high-level, \emph{out-of-distribution} examples (called canaries) are inserted into a training corpus, and a model trained on this corpus is then evaluated using various techniques to measure the extent to which the model has \emph{memorized} the canaries.
Since datasets in FL are inherently partitioned according to users, we adapt this framework to the FL regime by
 introducing two parameters to control the presence of a  canary in such settings.  An illustration of our federated secret sharer framework is shown in Figure~\ref{fig:fss}.
 Given a canary with parameters $p_u$ and $p_e$, we let $p_u$ be the  probability with which each user in a dataset is selected to be a secret sharer of the canary (Figure~\ref{fig:p_u}), whereas $p_e$ denotes the probability with which each example in such a secret sharer's data is replaced by the canary (Figure~\ref{fig:p_e}).

Our empirical evaluations (Section~\ref{sec:expts}) for this paper demonstrate the following key contributions. First, we show clustering the training data according to users, (i.e., having heterogeneous users) has a significant effect in reducing unintended memorization. Note that such a clustering of the data happens by design in FL settings. Next, given data clustered according to users, we show that replacing the learning optimizer from SGD to Federated Averaging provides a further reduction in such memorization. Lastly, we demonstrate that training in FL with a strong user-level DP guarantee results in models that exhibit the least amount of unintended memorization.

\mypar{Organization of the paper}
We provide a formal definition of differential privacy in Section~\ref{sec:dp}.
In Section~\ref{sec:vs}, we identify the main components separating central learning from the canonical setting of FL.
Section~\ref{sec:expts} contains the results of our empirical evaluation.
We state the conclusions of this work in Section~\ref{sec:conc}.

\subsection{Related Work}
\label{sec:rel}

There is a wide variety of work demonstrating unexpeted information leakage from datasets in unexpected ways: \cite{DN03} design a general \emph{reconstruction attack}
whereas there are other works (e.g., \cite{Homer08, Sank2009, BUV14, DSSUV15, mem-inf1,MSDCS18}) that design \emph{membership inference attacks}.
Apart from \cite{CLKES18} (which this work builds upon), other works~\cite{SS19} have also studied memorization in generative text models.

The FL paradigm, which is a major focus of this work, has been used to train multiple production scale models~\cite{NWP18, fl_emoji, fl_ngram}. We refer the reader to \cite{FLO} which provides an excellent overview of the state-of-the-art in the field, along with a suite of interesting open problems.

This work also studies the effectiveness of a user-level DP guarantee in reducing unintended memorization.  While many works on DP focus on \emph{record-level} DP guarantees (which usually cannot be directly extended to strong user-level DP guarantees), recent works (e.g., \cite{MRTZ18, JTT18, DPFGAN19, TAM19}) have designed techniques tailored to user-level DP guarantees.

\section{Background on Differential Privacy}
 \label{sec:dp}

 To establish the notion of differential privacy \cite{DMNS06, ODO}, we first define neighboring datasets. We will refer to a pair of datasets $D,D'$ as neighbors if $D'$ can be obtained by the addition or removal of all the examples associated with one user from $D$.

\begin{defn}[Differential privacy \cite{DMNS06, ODO}] A randomized  algorithm $\cA$ is $(\eps,\delta)$-differentially private if, for any pair of neighboring datasets $D$ and $D'$, and for all events $\cS$ in the output range of $\cA$, we have
$$\Pr[\cA(D)\in \cS] \leq e^{\eps} \cdot \Pr[\cA(D')\in \cS] +\delta$$
where the probability is taken over the random coins of $\cA$.
\label{def:diffP}
\end{defn}

For meaningful privacy guarantees, $\epsilon$ is assumed to be a small constant, and $\delta \ll 1/|D|$.

To train models with DP guarantees, we follow the variant of DP Federated Averaging (DP-FedAvg) \cite{MRTZ18} used in \cite{DPFGAN19}, where the only change is sampling fixed-sized minibatches in each training round.\footnote{Due to a technical limitation of the simulation framework, our experiments use sampling with replacement instead of without replacement; this should have negligible impact on the metrics of the trained models.} We provide more details of this technique, its privacy analysis, and a pseudo-code for reference in \ifsupp Appendix~\ref{sec:dp_fa}. \else the supplementary material. \fi

\section{Contrasting Federated Learning with Central Learning}
\label{sec:vs}

Now, we take a deeper look at how the well-studied  central learning framework differs from the canonical setting of Federated Learning (FL)~\cite{FL1}. Specifically, we are interested in differences that might have an effect on unintended memorization. We identify three such  components\footnote{We do not discuss unbalanced datasets, i.e., the fact that users can have varying amounts of local data, since FedAvg deals with such imbalances by weighing each client update according to the size of its local data.}:
\begin{enumerate}
    \item \emph{Learning Technique:} In central learning, the model is updated via an SGD step on a minibatch of records. In the canonical setting of FL, a model update typically corresponds to Federated Averaging over a minibatch of users: an average of the differences between the current model and the model obtained after several SGD steps on the local data of a user.
    \item \emph{Data Processed per Update:} Central learning typically ingests data as \emph{records}. For instance, the data for training a language model in central learning could be stored a set of sentences. On the other hand, FL operates at the granularity of a \emph{user}, with each user having their own set of records locally. In our example, the data in an FL setting would be a set of users, with each user having their own set of sentences locally.
    Typically, the amount of data processed per model update in central learning is much smaller in comparison to FL.
    \item \emph{Independent and Identically Distributed (IID) Data:} To reduce variance in learning, the data in central learning is shuffled before training (and/or each update involves a randomly sampled minibatch).
    Thus, each minibatch can be estimated to be drawn IID from the data. However, the data in FL is naturally grouped according to heterogeneous users, resulting in non-IID data even though each minibatch of users may be randomly sampled.
\end{enumerate}

Unintended memorization typically corresponds to the model memorizing information pertaining to a specific individual, or a very small group of individuals in the dataset. Given the above-stated differences between the two settings,  each of the three components could \emph{intuitively} play a significant role on such memorization.  For instance, it is conceivable that the higher amount of data processed per update in FL is better at concealing a targeted \emph{signal}, since each model update is averaged over effectively many more records than in central learning. Moreover, signals restricted to a small group of users may be encountered less frequently during FL training, since the data is clustered according to users. On the flip side, such heterogeneity (non-IIDness) of the data could result in a \emph{boosted} targeted signal whenever it is encountered in training. Additionally, since FedAvg involves several steps of SGD over user data in an update, the signal boost could be magnified due to factors such as ``client-drift"~\cite{SCAF}. Thus, it is unclear what effect each (or, any combination) of these differing components may have on such memorization in models trained via FL.

\section{Empirical Evaluation}
\label{sec:expts}
\subsection{Experimental Setup}
\label{sec:setup}
\mypar{Model Architecture} We use a word-level language model based on a recurrent neural network. Our model architecture mirrors the one used in \cite{NWP18}. The model comprises a CIFG-LSTM~\cite{cifg} with shared weights between the input embedding layer and the output projection layer. A vocabulary of size $\approx$10K is used for both input and output vocabularies. The input embedding dimension and the output projection size are set to 96, and a single-layer CIFG cell with 670 units is used as the recurrent layer. The overall number of parameters in the model is $\approx$1.3 M.

\mypar{Dataset Construction} We create a modified version of the Stack Overflow questions and answers dataset \cite{so_data} hosted by TensorFlow Federated \cite{ingerman2019tff}. The original dataset is keyed by a unique ID associated with the user who typed a question/answer. This ID can be used to separate the dataset into users for Federated Learning.
To compare the extent of memorization among different canary configurations in our federated secret sharer framework (Figure~\ref{fig:fss}), we modify the dataset to have balanced users as follows. First, we discard all users that have fewer than $2$K words. For the remaining users, we split the data associated with each user into (multiple) ``balanced" users having $\approx2$K words each. We also discard any ``remainder data'', i.e, if a user originally has 4030 words, we create 2 balanced users having $2$K words each and discard the remaining 30 words. The modified version of the dataset has $\approx392$K users, each with $\approx2$K words, and a total of $\approx93$K examples. For an IID version of this dataset, we randomly shuffle all the records in the dataset, and create \emph{synthetic} users having $\approx2$K words assigned sequentially from the shuffled records.

\mypar{Canary Construction} Since our model is a word-level language model, we follow the methodology used by \cite{CLKES18} for their experiments with the GMail Smart Compose model~\cite{smartcomp}. We opt for inserting 5-word canaries instead of 7-word canaries because our model is smaller\footnote{Our model is smaller because it is designed to run on mobile devices \cite{NWP18}.} than the Smart Compose model, and it is not as efficient at encoding longer contexts. Our canaries are constructed by choosing each word uniformly at random from the $10$K model vocabulary. To measure \emph{unintended} memorization, it is important to ensure that canaries comprise of out-of-distribution phrases. Randomly sampled 5-word canaries produces phrases that are highly unlikely to be in-distribution. For instance, our inserted canaries consist of phrases like ``qualifier winded alike configs txt", ``gentle originally saml likewise notified", ``dns scoring investigate compact auto", etc.

\mypar{Canary Insertion} Using our federated secret sharer framework (Figure~\ref{fig:fss}), we insert various canaries into the dataset.
Recall that each canary is parameterized by two parameters: a user-selection probability denoted by $p_u$, and an example-replacement probability denoted by $p_e$. We use Poisson sampling for both user-selection and example-replacement.
We insert canaries with configurations in the cross product of $p_u \in \{1/5\text{K},  3/50\text{K}, 1/50\text{K}\}$ and $p_e \in \{1\%, 10\%, 100\%  \}$.
 To limit the variance in our measurements, we insert 10 different canaries for each $(p_u, p_e)$ configuration.
These parameters result in the insertion of 90 different canaries, each shared among at least 4 users and at most 103 users in the dataset, with their overall insertion frequencies ranging from 13 to $24.5$K.

\mypar{Evaluation Methods} We use two methods of evaluation to understand the extent of unintended memorization exhibited by a model.
\begin{enumerate}
    \item \textbf{Random Sampling (RS) \cite{CLKES18}} It is a sampling-based method that captures how strongly the model favors the canary as compared to random chance. More specifically, we first define the log-perplexity of a model $\theta$ on a sequence $s = s_1, \ldots, s_n$ given context $p$ as $P_\theta(s | p) = \sum\limits_{i=1}^n \left(- \log \Pr\limits_\theta{(s_i|p,s_1, \ldots, s_{i-1})} \right)$. Next, this method requires i) a model $\theta$, ii) an inserted canary $c=(p|s)$ where $p$ is a prefix and $s$ is the remaining sequence, and iii) a set $R$ that consists of $|s|$-length sequences with each word sampled uniformly at random from the vocabulary. Now, the rank of the canary $c$ can be defined as $\text{rank}_\theta(c; R) = \left|
\{r' \in R: P_\theta(r' | p) \leq P_\theta(s | p)   \}\right|$. Using this method, we consider a canary $c$ as ``memorized" by a model $\theta$ if given a set $R$, we get that $\text{rank}_\theta(c; R)=1$.
For our experiments, we consider the size of $R$ to be 2M.
\item \textbf{Beam Search (BS)} Given a prefix, and the total length of the phrase to be extracted, this method conducts a greedy beam search on a model. As a result, this method functions without the knowledge of the whole canary.
Using this method, we consider a canary as ``memorized" by a model if given a prefix, the canary is the most-likely continuation.  For our experiments, we use a beam search width of 5 for this method.
\end{enumerate}
We perform measurements for both the methods using a prefix length of 1.
It is important to observe that since the BS method can function without complete knowledge of the inserted canary, it requires \emph{strictly} less information than the RS method.
Thus,
if a canary does not show up as memorized via the RS method, it will almost certainly not show up as memorized via the BS method.\footnote{Since beam search is essentially a greedy algorithm, it is possible to construct contrived examples where the BS method classifies a phrase as memorized whereas the RS method does not. However, given the typical log-perplexities of trained models on random phrases, the chance of observing such cases is very small.}

Along with evaluating for unintended memorization, we measure the utility of a model with accuracy (Recall@1) and perplexity on the test partition of the unmodified StackOverflow dataset.

\subsection{Empirical Results}
\label{sec:emp}
We present the results of our experiments on evaluating unintended memorization under different training regimes ranging from the well-studied central learning setting to canonical FL. For all our experiments that use SGD, we train models for $37.5$M steps, whereas we train for 8000 rounds for the experiments using FedAvg. For the largest minibatch sizes used in both the settings (256 records for SGD, and 5000 users for FedAvg), these checkpoints correspond to training for 100 epochs.
 Table~\ref{table:results_same_ckpt} shows the number of canaries (out of 90) that show up as memorized via both the RS and BS techniques.
 For both the methods, we also provide the lowest canary configuration (ordered by expected insertion frequency, and then by expected number of users sharing the canary) in terms of $(p_u, p_e)$ that had at least one of the ten inserted canaries as memorized.
 It is important to note that the utility of all the evaluated models is similar; the Recall@1 varies by less than 1\% across all the models. We conduct experiments for the following training regimes.
\begin{enumerate}
    \item \textbf{Central Learning:} First, we present the results for the central training regime where data is accessed as records, shuffled (IID), and SGD is used as the update step. This setting is the closest to the one evaluated in prior work~\cite{CLKES18}. We start by keeping the minibatch size $b_r=32$ records with a tuned learning rate of 0.005, and we double the minibatch size until it is 256.\footnote{In \ifsupp Appendix~\ref{app:expts}, \else the supplementary material, \fi we provide results for the setting where we increase the learning rate with the minibatch size, and stop training at 10 epochs for every minibatch size. We see that the amount of unintended memorization is relatively unaffected, but the model utility significantly decreases with increasing minibatch size.}  For all the configurations, we observe that 52-54  canaries  show up as memorized via the RS method, and 42-45 via the BS method.
    \item \textbf{FedAvg in Central Learning Setup:} Next, we evaluate the regime where the learning technique of SGD in the setup for central learning is replaced by FedAvg. Since data is operated at the granularity of ``users" for FedAvg, we create \emph{synthetic users}, each containing 2000 words, from the shuffled dataset for SGD. As a consequence, in this experiment we effectively have FedAvg operating over IID users. For all the configurations, 65-69 canaries  show up as memorized via the RS method, and 56-58 via the BS method.
    \item \textbf{Non-IIDness in Central Learning:} In this experiment, we evaluate the effect of Non-IID data in the setup of central learning. Since data is accessed as records for SGD but the natural grouping in the dataset is by heterogeneous users, we run the steps of SGD sequentially through the user-grouped variant of the dataset in each epoch of training. Here, we observe that the grouping of the data according to users exhibits a significant reduction in the unintended memorization by a trained model: for all the configurations, we observe that 37-51 canaries  show up as memorized via the RS  method, and 19-39 via the BS method.
    \item \textbf{Federated Learning:} Now, we evaluate the standard setting of FL where data is grouped by users (Non-IID), and FedAvg is used as the update step. The amount of unintended memorization in this setting significantly drops compared to any of the three settings evaluated above: for all the configurations, we observe that only 19-26 canaries show up as memorized via the RS method, whereas at most 2 canaries are extracted via the BS method. Notice that for the RS method, the lowest canary configuration that gets memorized is when 1 in every 5K users shares the canary, and further, for the BS method, only when all of the data of such users is replaced by the canary.
\end{enumerate}

\begin{table}[ht]
    \centering
    \begin{tabular}{| c | c | c | c| c| c| c|}
     \hline
     Setting &  RS  & RS Lowest  &  BS  & BS Lowest  & Acc. & Perp.\\
     Data, Optimizer, Batch Size &   /90 &  $(p_u, p_e)$ &   /90 &  $(p_u, p_e)$ & \%  & \\
     \hline
     IID, SGD, $b_r = 32$ & 54 & 1/50K, 10\%  & 42 & 3/50K, 10\% & 24  & 62.2  \\
      IID, SGD, $b_r = 64$ &  54 & 1/50K, 10\%  & 42 & 3/50K, 10\%  & 24.1 & 61.5 \\
       IID, SGD, $b_r = 128$ & 52 & 1/50K, 10\%   & 45 & 3/50K, 10\% & 24 & 62 \\
        IID, SGD, $b_r = 256$ & 53 & 1/50K, 10\%  & 43 & 1/50K, 10\%&  24.1 & 61.1 \\
     \hline
     \hline
     IID, FedAvg, $b_u = 500$ & 66 & 1/50K, 10\% & 56 & 1/50K, 10\% & 24.6 & 57.5 \\
     IID, FedAvg, $b_u = 1000$ & 69 & 3/50K, 1\% & 58  & 1/50K, 10\%& 24.6  & 57.3 \\
     IID, FedAvg, $b_u = 2000$ & 67 & 3/50K, 1\%& 56 & 1/50K, 10\%&  24.6 & 57.4 \\
     IID, FedAvg, $b_u = 5000$ & 65 & 3/50K, 1\%& 58 & 1/50K, 10\%& 24.6 & 57.3 \\
     \hline
     \hline
     Non-IID, SGD, $b_r = 32$ & 37 & 1/50K, 10\% & 19 & 3/50K, 10\% & 23.7 & 64.3 \\
     Non-IID, SGD, $b_r = 64$ & 49 & 1/50K, 10\% & 36 & 1/50K, 10\%& 24.1  & 61.8 \\
     Non-IID, SGD, $b_r = 128$ & 48 & 1/50K, 10\% & 34 & 1/50K, 10\% & 24.1 & 61.5 \\
     Non-IID, SGD, $b_r = 256$ & 51 & 1/50K, 10\% & 39 & 3/50K, 10\%& 24.1 & 61.3 \\
     \hline
     \hline
     Non-IID, FedAvg, $b_u = 500$ & 21  & 1/5K, 1\%& 0 & - & 24.4  & 58.8 \\
     Non-IID, FedAvg, $b_u = 1000$ & 23 & 1/5K, 1\% & 1 & 1/5K, 100\%& 24.3  & 59.5 \\
     Non-IID, FedAvg, $b_u = 2000$ & 19  & 1/5K, 1\%& 1 & 1/5K, 100\% & 24.5  & 58.3 \\
     Non-IID, FedAvg, $b_u = 5000$ & 26 & 1/5K, 1\% &  2 & 1/5K, 100\%& 24.5  & 58.2 \\
     \hline
    \end{tabular}
    \caption{Results for the number of inserted canaries (out of 90) memorized via the RS and BS methods, the lowest (by insertion frequency) canary configurations that show up as memorized for each of the methods, and utility metrics for various models evaluated at $37.5$M steps when sampling records (SGD), and 8000 rounds when sampling users (FedAvg).}
    \label{table:results_same_ckpt}
    \end{table}

\subsubsection{Training with user-level DP}
Next, we evaluate the effect of training with a guarantee of DP \cite{DMNS06, ODO}, on the unintended memorization in a trained model.
 To be able to provide the strongest DP parameters while obtaining reasonable utility from the trained models, we conduct experiments only for our largest minibatch size of 5000 users per training round. The results are presented in Table~\ref{table:results_dp}.

\mypar{Only Clipping} To bound the contribution by each participating user, DP-FedAvg clips each user update before aggregating them from a minibatch of users and adding calibrated noise to guarantee DP. Following \cite{CLKES18}, we present results (rows containing ``FedAvg+Clip" in Table~\ref{table:results_dp}) for the case when user updates are clipped to a value of 0.2, but no noise is added. This results in an $(\infty,\delta)$-DP guarantee for any $\delta \in (0, 1)$. This experiment helps us observe the effect of only clipping on the unintended memorization exhibited by trained models.
With IID data, for the setting evaluated in Section~\ref{sec:emp} (8000 rounds, i.e., 100 epochs for minibatch size of 5000 users),
we observe that the RS method extracts 58 canaries with clipping, which is 7 fewer canaries when compared to without clipping. The BS method extracts 49, which is 9 fewer than without clipping. If we pick a model after training for 10 epochs, the amount of memorization significantly reduces to 28 canaries via the RS method, and 18 via the BS method, but the utility of the resulting model is also significantly reduced. For Non-IID data, we observe a similar trend but it is more pronounced: the RS method extracts 11 canaries with clipping, which is 15 fewer canaries when compared to without clipping, whereas the BS method is not able to extract any of the inserted canaries. For the measurement at 10 epochs of training, both of our methods are able to extract no canary, but the utility of the model is reduced as well for this case.
\begin{table}[ht]
    \centering
    \begin{tabular}{| c | c | c | c| c| c| c|}
     \hline
     Setting ($b_u = 5000$) &  RS  & RS Lowest  &  BS  & BS Lowest  & Acc. & Perp.\\
     Data, Optimizer, Epochs &   /90 &  $(p_u, p_e)$ &   /90 &  $(p_u, p_e)$ & \%  & \\
     \hline
     IID, FedAvg, 100 & 65 & 3/50K, 1\%& 58 & 1/50K, 10\%& 24.6 & 57.3 \\
     IID, FedAvg+Clip, 100 & 58 & 1/50K, 10\%  & 49& 1/50K, 10\% & 24.2  & 60  \\
     IID, DP-FedAvg, 100  & 48  & 1/50K, 10\%& 42 & 3/50K, 10\%& 23.9  & 63 \\
     \hline
     IID, FedAvg+Clip, 10  & 28 & 1/50K, 100\%& 18  & 3/50K, 100\%& 21.9  & 83.2 \\
     IID, DP-FedAvg, 10 & 26  & 1/50K, 100\%& 18 & 3/50K, 100\%& 21.8  & 84.2 \\
     \hline
     \hline
     Non-IID, FedAvg, 100 & 26 & 1/5K, 1\% &  2 & 1/5K, 100\%& 24.5  & 58.2 \\
     Non-IID, FedAvg+Clip, 100 & 11 & 1/5K, 10\% & 0 & -& 24  & 61.5 \\
     Non-IID, DP-FedAvg, 100 & 12 & 1/5K, 10\% & 0& - & 23.3  &  68.5 \\
     \hline
     Non-IID, FedAvg+Clip, 10 & 0& - & 0& - & 20.8 & 95.9 \\
     Non-IID, DP-FedAvg, 10 &  0 & -& 0 & - & 20.7  & 97.1 \\

     \hline
     \end{tabular}
    \caption{Unintended memorization, lowest (by insertion frequency) canary configuration memorized, and utility for models trained with Clipping/DP and 5000 users/round. For 100 epochs, the models trained with DP-FedAvg satisfy $(18.8, 10^{-7})$-DP, and $(5.6, 10^{-7})$-DP for 10 epochs.}
    \label{table:results_dp}
    \end{table}

\mypar{DP} For obtaining a strong user-level DP guarantee, we add Gaussian noise with a noise multiplier of 1 to the per-round aggregate of clipped updates, which results in $(18.8, 10^{-7})$-DP for 100 epochs of training. We also provide measurements for 10 epochs of training, which results in a stronger $(5.6,10^{-7})$-DP guarantee. Compared to the results with only clipping, we observe a significant drop in memorization for the setting with the highest amount of memorization with only clipping, i.e., the setting with IID data evaluated at 100 epochs.

\subsection{Discussion}
\label{sec:disc}
\begin{enumerate}
    \item \textbf{Effect of clustering data according to users:} The results from our experiments strongly indicate that clustering data according to users significantly reduces unintended memorization. This is evident by considering the measurements in Table~\ref{table:results_same_ckpt} in pairs where the only differing component among them is whether the data is IID or not. The number of epochs taken over the dataset to train the models on which we measure memorization is the same for any particular minibatch size, irrespective of whether the data is IID. Thus, the number of times the inserted canaries were encountered during training is also comparable. However, the amount of memorization observed is always lower when the data is Non-IID. This effect is more pronounced in the settings where FedAvg is used as the training method. For instance, for a minibatch size $b_u=500$ users, training with FedAvg on IID data results in 66 canaries showing up as memorized via the RS method, and 56 via the BS method. However, the same configuration on Non-IID data results in the RS method classifying only 21 canaries as memorized, and the BS method not being able to extract any of the inserted canaries even after 8000 rounds of training. In addition to the data being clustered, the inserted canaries are clustered as well, which seems to play a crucial role in reducing such memorization. It is important to note that the utility of the models is similar for non-IID data even as the memorization drops significantly.\footnote{It might appear from Table~\ref{table:results_same_ckpt} that using FedAvg consistently provides better utility than SGD. However, SGD is sensitive to the tuning of the learning rate parameter \cite{FL1}, and with further fine-tuning we expect SGD to provide the same utility as FedAvg.}
    \item \textbf{Effect of varying data processed per update:} From the results presented in Table~\ref{table:results_same_ckpt}, fixing the optimizer to be from SGD/FedAvg and the data to be IID/non-IID, we do not observe any significant effect of varying the batch size, i.e., the data processed per update, on the unintended memorization of a model.
    \item \textbf{Effect of training heterogeneous user data with FedAvg and \emph{larger} minibatches:} The smallest minibatch size used for our experiments using FedAvg is 500 users per round, and as each user contains $\approx250$ records, the \emph{effective} minibatch size for this setting is $\approx125$K records. In comparison, the largest minibatch size we are able to conduct training for our experiments using SGD is 512 records. Focusing our attention on the results in Table~\ref{table:results_same_ckpt} using Non-IID data, we find that using FedAvg as an optimizer\footnote{We also conducted experiments using Momentum~\cite{mom} and Adam~\cite{adam} optimizers, but did not observe a strong effect on reducing memorization while maintaining comparable utility.}, and consequently having larger effective minibatches per round training causes a significant reduction in unintended memorization when compared to training with SGD.\footnote{Looking at the same set of results for IID data, the trend seems to be moving in the direction of increasing memorization. However, since the magnitude of the effect is much smaller, we deem that further investigation is required for this case, which leave for future work.} Even from the lowest canary configurations memorized in both the RS and the BS methods, it can be seen that this setting can tolerate a larger frequency of inserted canaries and/or canaries shared among a higher proportion of users.
    \item \textbf{Effect of adding Differential Privacy to FL training:} Some of our inserted canaries consist of being shared by as many as 103 users (with more than $24.5$K occurrences in the training data). By definition, a user-level DP guarantee is intended to be resilient to changes in the trained models w.r.t. any one user's data. Moreover, it is clear from the results in Table~\ref{table:results_same_ckpt} that the models trained in the FL regime exhibit the least amount of unintended memorization. In spite of these, we observe that training with a user-level DP guarantee for our largest minibatch size of 5000 users results in a significant further reduction in such memorization. With Non-IID data (which is what we expect naturally in a real-world implementation of FL), both of our RS/BS methods fail to extract any of the inserted canaries at 10 epochs of training. Here, we obtain a guarantee of $(5.6, 10^{-7})$-DP). Even at 100 epochs of training, where the guarantee becomes $(18.8, 10^{-7})$-DP, only the RS method is able to classify 12 canaries as memorized.  For the lowest canary configuration memorized in the RS method, notice that this setting can tolerate a magnitude larger frequency insertion of canaries than the setting without DP-FedAvg.
    It may not be surprising to believe the argument that as more privacy-preserving noise is added during training, less unintended memorization takes place. However, our results are noteworthy as, in spite of our DP models exhibiting the least amount of unintended memorization, they also provide a utility comparable to that of the models trained in Section~\ref{sec:emp}, along with providing a strong user-level guarantee of $(18.8, 10^{-7})$-DP.
\end{enumerate}

\section{Conclusion}
\label{sec:conc}
In this work, we conduct a formal study to understand the effect of the different components of Federated Learning, on the unintended memorization in trained models, as compared to the well-studied central learning. From our results, we observe that the components of FL exhibit a synergy in reducing such memorization. In particular, user-based heterogeneity of data (which occurs as a natural consequence in the FL setting) has a significant effect in the reduction, and training using Federated Averaging reduces it further. Moreover, we observe the least amount of memorization in the models where we train in FL with strong user-level differential privacy guarantees.

Recent work~\cite{SCAF} has shown that, in general, such heterogeneity in the training data can result in a slower and unstable convergence due to factors such as ``client-drift". For all of the experiments with non-IID data, we observe that the utility of the trained models is comparable to those trained on IID data, and we leave further exploration into why client-drift may not play a significant role in our experiments for future work.
Lastly, the secret-sharer line of methods for measuring unintended memorization operate at the granularity of a record. For future work, it will be interesting to design stronger attacks targeting data at the granularity of a user, and measure the resilience of models trained via FL, against such memorization.

\section*{Acknowledgements}
\label{sec:acks}
The authors would like to thank Mingqing Chen, and Andrew Hard for their helpful comments towards improving the paper.

\bibliographystyle{alpha}
\bibliography{references}
\newpage
\appendix
\section{DP Federated Averaging with Fixed-size Rounds}
 \label{sec:dp_fa}

    We now present the technique used to train our DP NWP model. It closely follows the DP-FedAvg technique in \cite{MRTZ18}, in that per-user updates are clipped to have a bounded $L_2$ norm, and calibrated Gaussian noise is added to the weighted average update to be used for computing the model to be sent in the next round. A slight difference between the DP-FedAvg algorithm in \cite{MRTZ18} and our approach is the way in which client devices are sampled to participate in a given federated round of computation. DP-FedAvg uses Poisson sampling, where for each round, each user is selected independently with a fixed probability. In this work (also, following \cite{DPFGAN19}), we instead use fixed-size federated rounds, where a fixed number of users is randomly sampled to participate in each round. For reference, we provide a pseudo-code for the technique in  Algorithm~\ref{alg.fedavg}.

\begin{figure}[ht]
\begin{small}
\centering
\rule{\textwidth}{0.4pt}
\begin{minipage}[t]{0.52\textwidth}
\begin{center}
\begin{algorithmic}
\SUB{Main training loop:}
   \STATE \emph{parameters:}  round participation fraction $q \in (0, 1]$,  total user population $N \in \mathbb{N}$, noise scale $z \in {\R}^+$, clip parameter $S \in {\R}^+$
   \STATE
   \STATE Initialize model $\theta^0$, moments accountant $\moments$
   \STATE Set $\sigma = \frac{\displaystyle z S}{\displaystyle qW}$
   \FOR{each round $t = 0, 1, 2, \dots$}
     \STATE $\sample^t \leftarrow$ (sample without replacement $q N$ users from population)
     \FOR{each user $k \in \sample^t$ \textbf{in parallel}}
       \STATE $\update^{t+1}_k \leftarrow \text{UserUpdate}(k, \theta^t)$
     \ENDFOR
     \STATE $\update^{t+1} = \frac{\displaystyle 1}{\displaystyle qN} \sum\limits_{k\in \sample^t} \update^{t+1}_k$
     \STATE $\theta^{t+1} \leftarrow \theta^t + \update^{t+1} + \Normal(0, I\sigma^2)$
     \STATE $\moments$.\texttt{accum\_priv\_spending}($z$)
   \ENDFOR
   \STATE print $\moments$.\texttt{get\_privacy\_spent}$()$
\end{algorithmic}
\end{center}
\vfill
\end{minipage}
\hfill
\begin{minipage}[t]{0.45\textwidth}
\begin{center}
\begin{algorithmic}
 \SUB{UserUpdate($k, \theta^0$):}
  \STATE \emph{parameters:}  number of local epochs $\lepochs \in \mathbb{N}$, batch size $\lbs \in \mathbb{N}$, learning rate $\eta \in {\R}^+$, clip parameter $S \in {\R}^+$, loss function $\loss(\theta; b)$
  \STATE
  \STATE $\theta \leftarrow \theta^0$
  \FOR{each local epoch $i$ from $1$ to $\lepochs$}
    \STATE $\mathcal{B} \leftarrow$ ($k$'s data split into size $\lbs$ batches)
    \FOR{each batch $b \in \mathcal{B}$}
      \STATE $\theta \leftarrow \theta - \eta \grad \loss(\theta; b)$
    \ENDFOR
 \ENDFOR
 \STATE $\update = \theta - \theta^0$
 \STATE return update $\update_k = \update \cdot \min\left(1, \frac{S}{\norm{\update}}\right)$ \hfill \emph{// Clip}

\end{algorithmic}
\end{center}
\vfill
\end{minipage}
\end{small}
\rule{\textwidth}{0.4pt}

\mycaptionof{algorithm}{DP-FedAvg with fixed-size federated rounds, used to train our DP NWP model.}\label{alg.fedavg}

\end{figure}

\paragraph{Privacy analysis: }
Following the analysis of this technique in \cite{DPFGAN19}, we obtain our DP guarantees by using the following:
\begin{enumerate}
    \item the analytical moments accountant \cite{WBK19} to obtain the R\'enyi differential privacy (RDP) guarantee for a federated round of computation that is based on the subsampled Gaussian mechanism,
    \item Proposition 1~\cite{mir17} for computing the RDP guarantee of the composition involving all the rounds, and
    \item Proposition 3~\cite{mir17} to obtain a DP guarantee from the composed RDP guarantee.
\end{enumerate}

\section{Additional Empirical Evaluation}
\label{app:expts}
In this section, we present the results of our additional empirical evaluation that was omitted from the main body.

\mypar{Using Different Optimizers} In Table~\ref{table:results_same_epoch_diff_opts}, we provide the results for using different optimizers like Momentum~\cite{mom} and Adam~\cite{adam} for training. We conduct experiments using only the smallest batch size in both the granularities (32 records, or 500 users). For Momentum, we set the momentum parameter to $0.9$, and for Adam, we set the learning rate to $10^{-4}$. First, we observe that using Momentum increases the observed unintended memorization but has a similar utility as SGD. On the other hand, we see that using Adam decreases such memorization, but the utility of the models is also noticeably reduced as compared to SGD.
We observe a similar trend when Adam is combined with FedAvg.

\begin{table}[ht]
\centering
\begin{tabular}{|c|c|c|c|c|c|c|}
\hline
Setting                       & RS  & RS Lowest    & BS  & BS Lowest    & Acc. & Perp. \\
Data, Optimizer, Batch Size   & /90 & $(p_u, p_e)$ & /90 & $(p_u, p_e)$ & \%   &       \\ \hline
IID, SGD, $b_r = 32$ & 54 & 1/50K, 10\%  & 42 & 3/50K, 10\% & 24  & 62.2  \\
IID, Momentum, $b_r = 32$         & 64  & 3/50K, 1\%  & 50  & 1/50K, 10\%  & 24.2 & 60.6    \\
IID, Adam, $b_r = 32$        & 56  & 1/50K, 10\%  & 42  & 3/50K, 10\%  & 22.7 & 70.8    \\
\hline
IID, FedAvg, $b_u = 500$      & 66  & 1/50K, 10\%  & 56  & 1/50K, 10\%  & 24.6 & 57.5  \\
IID, FedAvg+Adam, $b_u = 500$      & 60  & 1/50K, 10\%  & 46  & 1/50K, 10\%  & 23.7 & 63.7  \\\hline
Non-IID, SGD, $b_r = 32$ & 37 & 1/50K, 10\% & 19 & 3/50K, 10\% & 23.7 & 64.3 \\
Non-IID, Momentum, $b_r = 32$       & 48  & 3/50K, 1\%  & 36  & 1/50K, 10\%  & 24.3 & 59.9  \\
Non-IID, Adam, $b_r = 32$       & 21  & 1/50K, 10\%  & 13  & 1/50K, 10\%  & 21.5 & 84.1  \\
 \hline
Non-IID, FedAvg, $b_u = 500$  & 21  & 1/5K, 1\%    & 0   & -            & 24.4 & 58.8  \\
Non-IID, FedAvg+Adam, $b_u = 500$  & 8  & 1/5K, 10\%    & 0   & -            & 23.3 & 66.5  \\ \hline
\end{tabular}
\caption{Unintended memorization, lowest (by insertion frequency) canary configuration memorized, and utility metrics for models using different optimizers evaluated at $37.5$M steps when sampling records (e.g., SGD), and 8000 rounds when sampling users (e.g., FedAvg).}
    \label{table:results_same_epoch_diff_opts}
\end{table}

\mypar{Evaluating for Same Training Epochs} In Table~\ref{table:results_same_epoch}, we provide the results for evaluating models trained for the same number of epochs over the training data. For the runs using SGD, we start with a batch size of 32 records and a tuned learning rate of 0.005, and we increase the learning rate by $\approx\sqrt{2}$ for every $2$x increase in the batch size.
For all the experiments with FedAvg, we find that using a constant learning rate provides the best utility across the different batch sizes, and thus, we keep it fixed.
For the models trained with SGD, for both IID and non-IID data we observe that unintended memorization remains comparable for models trained with different batch sizes. However, we see a decrease in the utility as the batch size increases.  The decrease in utility is observed for models trained using FedAvg as well, but we also observe a significant drop in the such memorization when training is performed with at least 1000 users per round. Moreover, once the training involves at least 2000 users on non-IID data, both the RS and BS methods are unsuccessful in classifying any of the 90 inserted canaries as memorized.

\begin{table}[ht]
\centering
\begin{tabular}{|c|c|c|c|c|c|c|}
\hline
Setting                       & RS  & RS Lowest    & BS  & BS Lowest    & Acc. & Perp. \\
Data, Optimizer, Batch Size   & /90 & $(p_u, p_e)$ & /90 & $(p_u, p_e)$ & \%   &       \\ \hline
IID, SGD, $b_r = 32$          & 49  & 1/50K, 10\%  & 43  & 3/50K, 10\%  & 23.9 & 63    \\
IID, SGD, $b_r = 64$          & 51  & 1/50K, 10\%  & 39  & 3/50K, 10\%  & 23.5 & 65.1  \\
IID, SGD, $b_r = 128$         & 46  & 3/50K, 10\%  & 36  & 1/50K, 100\% & 23.2 & 67.6  \\
IID, SGD, $b_r = 256$         & 46  & 3/50K, 10\%  & 32  & 1/50K, 100\% & 23   & 69.7  \\ \hline
IID, FedAvg, $b_u = 500$      & 66  & 1/50K, 10\%  & 56  & 1/50K, 10\%  & 24.6 & 57.5  \\
IID, FedAvg, $b_u = 1000$     & 27  & 1/50K, 100\% & 18  & 3/50K, 100\% & 24.5 & 58    \\
IID, FedAvg, $b_u = 2000$     & 28  & 1/50K, 100\% & 18  & 3/50K, 100\% & 24.4 & 59.3  \\
IID, FedAvg, $b_u = 5000$     & 28  & 1/50K, 100\% & 18  & 3/50K, 100\% & 24   & 62.5  \\ \hline
Non-IID, SGD, $b_r = 32$      & 40  & 1/50K, 10\%  & 29  & 1/50K, 10\%  & 23.7 & 64.2  \\
Non-IID, SGD, $b_r = 64$      & 38  & 1/50K, 10\%  & 32  & 1/50K, 10\%  & 23.6 & 65.6  \\
Non-IID, SGD, $b_r = 128$     & 35  & 1/50K, 10\%  & 26  & 1/50K, 100\% & 23.2 & 68.2  \\
Non-IID, SGD, $b_r = 256$     & 40  & 3/50K, 10\%  & 31  & 1/50K, 100\% & 23   & 69.8  \\ \hline
Non-IID, FedAvg, $b_u = 500$  & 21  & 1/5K, 1\%    & 0   & -            & 24.4 & 58.8  \\
Non-IID, FedAvg, $b_u = 1000$ & 10  & 1/5K, 10\%   & 0   & -            & 24   & 61.2  \\
Non-IID, FedAvg, $b_u = 2000$ & 0   & -            & 0   & -            & 24   & 61.9  \\
Non-IID, FedAvg, $b_u = 5000$ & 0   & -            & 0   & -            & 23.3 & 67.9  \\ \hline
\end{tabular}
\caption{Results for the number of inserted canaries (out of 90) memorized via the RS and BS methods, the lowest (by insertion frequency) canary configurations that show up as memorized for each of the methods, and utility metrics for various models evaluated at $\approx$10 epochs of training.}
    \label{table:results_same_epoch}
\end{table}

\end{document}